\documentclass[letterpaper, paper,11pt]{AAS}		

\usepackage{bm}  
\usepackage{amsmath}
\usepackage{amssymb}  
\usepackage{subfigure}
\usepackage{booktabs}                    
\usepackage[colorlinks=true, pdfstartview=FitV, linkcolor=black, citecolor= black, urlcolor= black]{hyperref}
\usepackage{overcite}
\usepackage{footnpag}			      	
\usepackage{tikz}
\usepackage{xcolor}
\usetikzlibrary{arrows,shapes,positioning,fit,matrix,backgrounds,calc,decorations.pathreplacing,math}
\usepackage[capitalise,noabbrev]{cleveref}
\usepackage{algorithm}
\usepackage{algpseudocode}

\PaperNumber{25-860}

\begin{document}

\title{Multi-Phase Spacecraft Trajectory Optimization via Transformer-Based Reinforcement Learning}

\author{Amit Jain\thanks{Postdoctoral Associate, Aeronautics \& Astronautics, Massachusetts Institute of Technology, Cambridge, MA 02139.},
Victor Rodriguez-Fernandez\thanks{Associate Professor, Department of Computer Systems Engineering, Universidad Polit\'ecnica de Madrid, 28040 Madrid, Spain},
\ and Richard Linares\thanks{Associate Professor, Aeronautics \& Astronautics, Massachusetts Institute of Technology, Cambridge, MA 02139.}
}

\maketitle{} 		

\begin{abstract}
Autonomous spacecraft control for mission phases such as launch, ascent, stage separation, and orbit insertion remains a critical challenge due to the need for adaptive policies that generalize across dynamically distinct regimes. While reinforcement learning (RL) has shown promise in individual astrodynamics tasks, existing approaches often require separate policies for distinct mission phases, limiting adaptability and increasing operational complexity. This work introduces a transformer-based RL framework that unifies multi-phase trajectory optimization through a single policy architecture, leveraging the transformer's inherent capacity to model extended temporal contexts. Building on proximal policy optimization (PPO), our framework replaces conventional recurrent networks with a transformer encoder-decoder structure, enabling the agent to maintain coherent memory across mission phases spanning seconds to minutes during critical operations. By integrating a Gated Transformer-XL (GTrXL) architecture, the framework eliminates manual phase transitions while maintaining stability in control decisions. We validate our approach progressively: first demonstrating near-optimal performance on single-phase benchmarks (double integrator and Van der Pol oscillator), then extending to multiphase waypoint navigation variants, and finally tackling a complex multiphase rocket ascent problem that includes atmospheric flight, stage separation, and vacuum operations. Results demonstrate that the transformer-based framework not only matches analytical solutions in simple cases but also effectively learns coherent control policies across dynamically distinct regimes, establishing a foundation for scalable autonomous mission planning that reduces reliance on phase-specific controllers while maintaining compatibility with safety-critical verification protocols.

\end{abstract}

\section{Introduction}
Space exploration and satellite operations require sophisticated trajectory optimization and control methods that navigate complex gravitational environments while satisfying stringent constraints on fuel consumption, mission timelines, and safety. Traditionally, spacecraft trajectory planning and control rely on separate policies designed for specific mission phases such as orbit raising, station-keeping, rendezvous, and landing, each requiring specialized expertise and computational approaches. This segmentation introduces operational complexity and potential failure points during phase transitions and limits the adaptability of spacecraft systems to changing mission requirements or unexpected environmental conditions. 

The rapid expansion of space infrastructure with nearly 150,000 satellites projected in low Earth orbit by 2035 \cite{esa2025traffic}, demands revolutionary approaches to autonomous spacecraft guidance that transcend traditional mission-phase segmentation. Space missions typically decompose trajectory optimization into discrete phases (orbit insertion, station-keeping, rendezvous, and landing), creating operational vulnerabilities during regime transitions and limiting adaptability to dynamic space environments. This fragmentation had persisted since the Apollo era when computational constraints necessitated dividing complex missions into manageable segments with specialized controllers for each phase \cite{trimble2006mission, chilan2013automated}.

Traditional model-based approaches, rooted in optimal control theory \cite{betts2010practical}, have dominated spacecraft guidance for decades. These methods formulate trajectory optimization as constrained convex programming problems solved through sequential quadratic programming \cite{acikmese2007convex} or pseudospectral techniques \cite{garg2011direct,jain2023sparse,jain2025hamilton}. While providing mathematical optimality guarantees, they require precise knowledge of spacecraft dynamics and environmental parameters, an assumption frequently violated by unmodeled disturbances, uncertain mass properties, and complex multibody gravitational fields \cite{scott2018autonomous}. The 2024 demonstration of NASA's Artemis II mission highlighted these limitations when unexpected lunar gravitational anomalies required manual trajectory corrections \cite{nasa2024artemis}, underscoring the need for adaptive control systems that maintain coherence across mission phases.

Reinforcement learning (RL) emerged as a promising alternative for spacecraft guidance following the development of policy gradient methods such as proximal policy optimization (PPO) \cite{schulman2017proximal}. While Schulman's original work presented PPO as a general-purpose algorithm, the aerospace community quickly recognized its potential for autonomous spacecraft control. Linares et al. \cite{gaudet2020deep,gaudet2020adaptive} pioneered its application to spacecraft control problems, demonstrating convergence without prior knowledge of inertia properties. Kolosa and Elkins \cite{kolosa2019reinforcement, elkins2020neural} extended these capabilities to trajectory optimization, showing RL could discover fuel-efficient transfer orbits while handling environmental uncertainties. However, these implementations remained phase-specific with limited temporal context, restricting their applicability to multi-day orbital maneuvers \cite{carradori2024atmospheric}. 

The integration of transformer networks with reinforcement learning represents a quantum leap in spacecraft guidance capabilities. Originally developed for natural language processing\cite{vaswani2017attention}, transformers' self-attention mechanisms excel at modeling long-range dependencies in sequential data, making them ideally suited for spacecraft trajectory optimization across multiple phases. While initial applications of transformers to reinforcement learning faced training instabilities, Parisotto et al. \cite{parisotto2020stabilizing} addressed these challenges by introducing the Gated Transformer-XL (GTrXL) architecture. This breakthrough stabilized transformer training for sequential decision-making tasks, opening new possibilities for spacecraft control applications.

\begin{figure}[b!]
   \centering
   \includegraphics[width=0.6\linewidth]{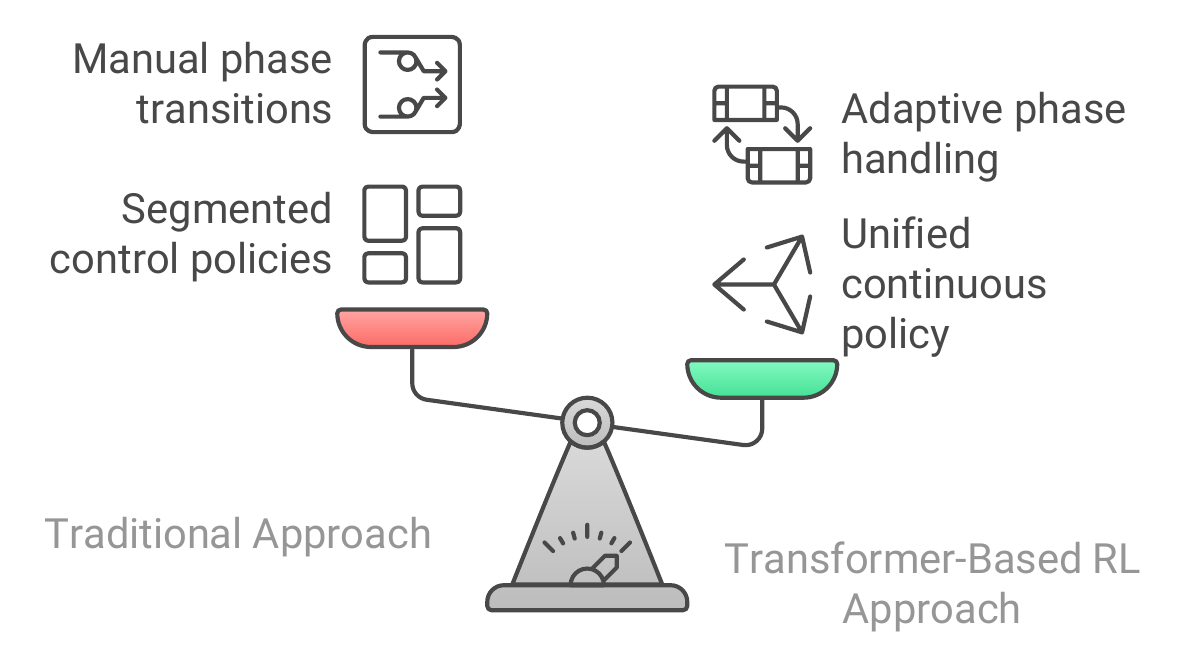}
   \caption{Comparison of traditional segmented spacecraft control approaches (left) versus our proposed transformer-based reinforcement learning framework (right).}
   \label{fig:comparison}
\end{figure}

Building on this foundation, researchers began applying transformer architectures to specific spacecraft mission phases. Federici et al. \cite{federici2022meta} demonstrated transformers' superiority over recurrent networks for handling mission-critical temporal dependencies during lunar landings. Their GTrXL implementation achieved 98.7\% landing success under combined navigation errors and thrust uncertainties by dynamically weighting sensor inputs through context-aware attention mechanisms. Similarly, Carradori et al. \cite{carradori2024atmospheric} adapted this framework for atmospheric rocket landings by introducing aerodynamic-aware attention masks that automatically prioritize relevant flow features during different descent phases.



Despite these advances, current transformer-RL implementations remain narrowly focused on single mission phases. Our work addresses this critical gap by developing a unified control framework capable of handling multi-phase spacecraft operations through a single adaptive policy. Building on the GTrXL foundation \cite{parisotto2020stabilizing}, we leverage its episodic memory mechanism to maintain temporal context across mission phases, capturing the transitions between dynamically distinct operational regimes. Our approach eliminates the need for explicit phase transitions through two key design choices: (1) a unified multi-objective reward function that smoothly transitions between phase-specific objectives without discrete switching, allowing natural progression from one mission segment to another, and (2) an augmented observation space that includes normalized time and system state information, enabling the transformer's attention mechanism to naturally adapt control strategies as dynamics evolve. The GTrXL's sliding memory window stores recent state-action histories, allowing the policy to recognize patterns indicating regime changes and adjust control accordingly without manual intervention or predefined switching logic.


Figure \ref{fig:comparison} illustrates the fundamental difference between traditional spacecraft control approaches and our proposed framework. Conventional methods (left) partition operations into discrete phases with separate controllers, creating inherent vulnerabilities at transition boundaries. Our transformer-based reinforcement learning architecture (right) creates a unified control policy that learns generalizable strategies that seamlessly adapt across multiple mission phases without explicit reformulation. The transformer's attention mechanisms enable it to identify relevant historical information, extract transferable control patterns, and automatically adjust control strategies based on the current mission context.

The remainder of this paper is organized as follows: The paper first formulates the optimal control problem as a Markov Decision Process suitable for reinforcement learning, then details the transformer-based reinforcement learning architecture, including the GTrXL model and PPO algorithm integration. The numerical simulation results follow a progressive validation approach: demonstrating near-optimal performance on single-phase benchmark problems (double integrator and Van der Pol oscillator) to establish the validity of the approach against known solutions. These benchmarks are then extended to multiphase variants with waypoint navigation, demonstrating the framework's ability to handle sequential objectives without architectural modifications. Finally, the method is applied to a complex multiphase rocket ascent problem, showcasing its capability to manage discontinuous dynamics, varying objectives, and long time horizons in a realistic aerospace application. The paper concludes with a discussion of limitations, future work, and broader implications for autonomous control systems in space applications.

\section{Problem Formulation}
\subsection{Optimal Control Problem}
The multiphase spacecraft trajectory optimization problems addressed in this work are formulated within the optimal control framework. Consider a general nonlinear dynamical system:

\begin{equation} 
\dot{\mathbf{x}}(t) = \mathbf{f}(\mathbf{x}(t), \mathbf{u}(t), t) 
\end{equation}

where $\mathbf{x}(t) \in \mathbb{R}^n$ is the state vector, $\mathbf{u}(t) \in \mathbb{R}^m$ is the control input, and $\mathbf{f}: \mathbb{R}^n \times \mathbb{R}^m \times \mathbb{R} \rightarrow \mathbb{R}^n$ represents the system dynamics.

The objective is to find a control policy that minimizes:

\begin{equation} 
J = \Phi(\mathbf{x}(t_f), t_f) + \int_{t_0}^{t_f} \mathcal{L}(\mathbf{x}(t), \mathbf{u}(t), t) dt 
\end{equation}

where $\Phi(\mathbf{x}(t_f), t_f)$ is the terminal cost, $\mathcal{L}(\mathbf{x}(t), \mathbf{u}(t), t)$ is the running cost, and $[t_0, t_f]$ is the time interval. The optimization is subject to:
\begin{align}
   \mathbf{x}(t_0) &= \mathbf{x}_0 \\
   \psi(\mathbf{x}(t_f), t_f) &= \mathbf{0}
\end{align}

where $\mathbf{x}_0$ is the initial state and $\psi(\mathbf{x}(t_f), t_f)$ represents terminal constraints.

Traditional methods for solving such problems include direct collocation, shooting methods, and pseudospectral approaches. However, multiphase spacecraft missions present unique challenges:
\begin{itemize}
\item \textbf{Discontinuous dynamics}: Mass drops at stage separations create discontinuities
\item \textbf{Phase-dependent objectives}: Each flight phase has distinct priorities
\item \textbf{Long time horizons}: Missions span vastly different dynamic regimes
\item \textbf{Model uncertainties}: Atmospheric variations and thrust dispersions affect performance
\end{itemize}

These challenges motivate the exploration of learning-based approaches that can adaptively handle varying dynamics and objectives across multiple phases without explicit reformulation at phase boundaries.

\subsection{Markov Decision Process Formulation}
To leverage reinforcement learning, the optimal control problem is reformulated as a Markov Decision Process (MDP). An MDP is defined by the tuple $(\mathcal{S}, \mathcal{A}, \mathcal{P}, \mathcal{R}, \gamma)$, where $\mathcal{S}$ represents the state space, $\mathcal{A}$ the action space, $\mathcal{P}: \mathcal{S} \times \mathcal{A} \times \mathcal{S} \rightarrow [0, 1]$ the transition probability function, $\mathcal{R}: \mathcal{S} \times \mathcal{A} \times \mathcal{S} \rightarrow \mathbb{R}$ the reward function, and $\gamma \in [0, 1]$ the discount factor.

For spacecraft control problems, the continuous dynamics are discretized using a fixed time step $\Delta t$:

\begin{equation}
\mathbf{x}_{k+1} = \mathbf{x}_k + \int_{t_k}^{t_k + \Delta t} \mathbf{f}(\mathbf{x}(t), \mathbf{u}(t), t) dt
\end{equation}

where $\mathbf{x}_k$ and $\mathbf{u}_k$ represent the state and control at time step $k$, respectively. This integration is performed using numerical methods such as Runge-Kutta schemes.

The specific MDP formulation consists of:

\begin{itemize}
\item \textbf{State space} $\mathcal{S}$: The system state vector including position, velocity, mass, and additional problem-specific variables. For multiphase problems, this includes implicit phase information through physical quantities (e.g., altitude, atmospheric density).

\item \textbf{Action space} $\mathcal{A}$: Control inputs such as thrust magnitude and direction, constrained by physical actuator limits. For discrete action spaces, thrust directions are discretized on a unit sphere.

\item \textbf{Transition dynamics}: Deterministic in our implementation, following the discretized system dynamics. Discontinuous transitions occur at phase boundaries (e.g., stage separation).

\item \textbf{Reward function} $\mathcal{R}$: Designed to encourage desired behavior across all phases:
\begin{equation}
r_k = -(\Delta t \cdot \mathcal{L}(\mathbf{x}_k, \mathbf{u}_k, t_k) + \mathcal{C}_{penalty}) + \mathcal{B}_{bonus}
\end{equation}
where $\mathcal{C}_{penalty}$ represents penalty terms for constraint violations, and $\mathcal{B}_{bonus}$ represents bonus terms for achieving phase-specific objectives. The reward structure unifies multiple objectives without explicit switching, transitioning smoothly from altitude gain and gravity turn execution in early phases to orbital element matching in terminal phases.

\item \textbf{Discount factor} $\gamma$: Set close to 1 (typically 0.99) to emphasize long-term mission success over short-term gains.
\end{itemize}

\section{Transformer-Based Reinforcement Learning Architecture}
Traditional neural network approaches for control, including feed-forward networks and recurrent neural networks (RNNs), face limitations in multiphase problems. Feed-forward networks process only current states, lacking temporal context crucial for recognizing phase transitions. RNNs maintain memory but suffer from vanishing gradients over long sequences and compress all history into fixed-size hidden states, limiting their effectiveness for missions spanning multiple operational regimes. Transformer architectures \cite{vaswani2017attention} address these limitations through self-attention mechanisms that can directly access any part of the input sequence. This capability is particularly valuable for multiphase control where decisions depend on events occurring at various timescales, from immediate state feedback to mission-elapsed-time awareness.

\subsection{Architecture Overview}
The proposed actor-critic architecture combines reinforcement learning with a Gated Transformer-XL (GTrXL) memory system \cite{parisotto2020stabilizing}. As illustrated in Figure~\ref{fig:transformer-ppo-architecture}, the system comprises four main components:

\begin{enumerate}
\item \textbf{Observation Encoder}: Transforms raw state observations into high-dimensional embedding vectors suitable for transformer processing
\item \textbf{Gated Transformer-XL Module}: Processes embeddings with temporal context via multi-head attention mechanisms
\item \textbf{Policy Head}: Outputs action distributions for control decisions
\item \textbf{Value Head}: Estimates state values for advantage computation
\end{enumerate}

\subsection{Gated Transformer-XL Implementation}
The GTrXL architecture modifies standard transformers for reinforcement learning stability. Unlike RNNs that compress history into fixed-size states, GTrXL maintains a sliding window of past embeddings (typically 32-128 time steps) accessible via attention. Key modifications for RL stability include:

\textbf{Gating Mechanism:} Standard residual connections $y = x + f(x)$ cause optimization instabilities in RL. GTrXL employs learnable gates:
\begin{equation} 
\text{gate}(x, y) = \lambda \odot x + (1 - \lambda) \odot y, \quad \lambda = \sigma(W_g [x, y] + b_g)
\end{equation}
with positive-initialized bias $b_g$ favoring identity mapping during early training.

\textbf{Transformer Blocks:} Each block processes inputs through gated attention and feed-forward layers:
\begin{align} 
h_1 &= \text{gate}(x, \text{LayerNorm}(x + \text{MultiHead}(x))) \\
h_2 &= \text{gate}(h_1, \text{LayerNorm}(h_1 + \text{MLP}(h_1))) 
\end{align}

\textbf{Relative Positional Encoding:} To capture when events occurred, attention scores incorporate relative positions:
\begin{equation} 
A_{i,j} = \frac{(x_i W^Q)(x_j W^K + R_{i-j})^T}{\sqrt{d_k}} 
\end{equation}
where $R_{i-j}$ encodes the relative distance between positions, enabling phase transition recognition from temporal patterns.

\subsection{Integration with Proximal Policy Optimization}
The transformer architecture is trained using Proximal Policy Optimization (PPO) \cite{schulman2017proximal}, ensuring stable policy updates through a clipped surrogate objective:
\begin{equation} 
L^{CLIP}(\theta) = \hat{\mathbb{E}}_t\left[\min\left(r_t(\theta)\hat{A}_t, \text{clip}(r_t(\theta), 1-\epsilon, 1+\epsilon)\hat{A}_t\right)\right] 
\end{equation}

where $r_t(\theta) = \frac{\pi_\theta(a_t|s_t)}{\pi_{\theta_{old}}(a_t|s_t)}$ is the probability ratio and $\hat{A}_t$ is the advantage estimate computed using Generalized Advantage Estimation (GAE):
\begin{equation}
\hat{A}_t = \sum_{l=0}^{\infty} (\gamma\lambda)^l \delta_{t+l}
\end{equation}
where $\delta_t = r_t + \gamma V(s_{t+1}) - V(s_t)$ is the temporal difference error.

\subsection{Training Procedure}

The proposed training procedure addresses multiphase challenges through careful memory management and temporal coherence preservation. Algorithm~\ref{alg:gtrxl-ppo} presents the complete process.

\begin{algorithm}[htb!]
\caption{GTrXL-PPO Training for Multiphase Control}
\label{alg:gtrxl-ppo}
\begin{algorithmic}[1]
\Require Iterations $N$, environments $E$, memory length $L_{mem}$, PPO epochs $K$
\State Initialize policy $\pi_\theta$, value function $V_\phi$, memory encoder $\psi$
\For{iteration $i = 1$ to $N$}
    \State Reset buffers: $\mathcal{D} \leftarrow \emptyset$, $\mathcal{M} \leftarrow \emptyset$
    \For{environment $e = 1$ to $E$ \textbf{in parallel}}
        \State $m_0^e \sim \mathcal{N}(0, \sigma^2 I)$, $s_0^e \leftarrow$ Reset environment
        \For{$t = 0$ to termination}
            \State $h_t^e \leftarrow \psi(s_t^e, m_t^e)$ \Comment{Encode with memory}
            \State $a_t^e \sim \pi_\theta(h_t^e)$, $v_t^e \leftarrow V_\phi(h_t^e)$
            \State Execute $a_t^e$: $s_{t+1}^e, r_t^e \leftarrow \text{Env}(a_t^e)$
            \State Update memory: $m_{t+1}^e \leftarrow \text{GTrXL}(m_t^e, s_t^e, a_t^e)$
            \State Store $(s_t^e, a_t^e, r_t^e, v_t^e)$ in $\mathcal{D}$, memory in $\mathcal{M}$
        \EndFor
    \EndFor
    \State Compute GAE advantages over trajectories
    \For{epoch $j = 1$ to $K$}
        \For{mini-batch $\mathcal{B} \subset \mathcal{D}$ preserving sequences}
            \State Retrieve memory contexts from $\mathcal{M}$
            \State Update $\theta, \phi$ via $L^{CLIP} + c_1 L^{VF} - c_2 H[\pi_\theta]$
        \EndFor
    \EndFor
\EndFor
\end{algorithmic}
\end{algorithm}

Key implementation details: (1) Memory states $m_t^e \in \mathbb{R}^{L_{mem} \times d}$ are initialized with small noise to encourage exploration. (2) Mini-batches preserve temporal structure through sequential grouping and overlapping windows. (3) Memory-conditioned value functions $V_\phi(s_t, m_t)$ enable accurate long-term predictions across phase boundaries.

\begin{figure}[htb!]
   \centering
   \begin{tikzpicture}[
   node distance=5cm,
   box/.style={draw, rounded corners, fill=blue!10, minimum width=3cm, minimum height=1.2cm, align=center, font=\footnotesize},
   memorybox/.style={draw, rounded corners, fill=green!10, minimum width=6cm, minimum height=3.5cm, align=center, font=\footnotesize},
   connector/.style={->, >=stealth, thick, line width=0.8pt},
   dashedconnector/.style={->, >=stealth, thick, dashed, color=green!50!black, line width=0.8pt},
   dashedbox/.style={draw, dashed, color=blue!50, thick, rounded corners},
   every text node part/.style={align=center},
   scale=0.6
   ]
   
   \node[box, minimum height=3cm, fill=gray!20] (env) {\textbf{Environment} \\ (Control System)};
   \node[below=0.1 of env, text width=3cm, font=\scriptsize] (envdesc) {Double Integrator\\Van der Pol Oscillator\\Rocket Ascent};
   
   \node[above=0.2cm of env, font=\scriptsize, color=blue!70, anchor=south] {Phase $\phi \in \{1,...,N\}$};
   
   \node[box, right=2cm of env, minimum width=3.5cm, minimum height=3cm, fill=blue!5] (model) {\textbf{Actor-Critic} \\ \textbf{Model}};
   
   \node[box, above=1.4cm of model, minimum width=3cm, fill=blue!20] (encoder) {\textbf{Observation Encoder}\\$\mathbf{h}_t = f_\theta(\mathbf{o}_t)$};
   
   \node[box, below left=2cm and 0.7cm of model, minimum width=3cm, fill=purple!10] (policy) {\textbf{Policy Head}\\$\pi_\theta(\cdot|s_t,m_t)$};
   \node[box, below right=2cm and 0.7cm of model, minimum width=3cm, fill=green!20] (value) {\textbf{Value Head}\\$V_\phi(s_t,m_t)$};
   
   \node[below=0.2cm of policy, font=\scriptsize, anchor=north] {$\mathbf{a}_t \sim \pi_\theta$};
   
   \node[memorybox, right=2cm of model, fill=green!5, minimum width=5cm, minimum height=3cm, text depth=2cm] (memory) {\textbf{Gated Transformer-XL}};
   
   \node[box, below=4.8cm of model, minimum width=3cm, fill=red!5] (trainer) {\textbf{PPO Trainer}\\$L^{CLIP} + c_1 L^{VF} - c_2 H[\pi]$};
   
   \node[dashedbox, fit=(encoder)(model)(memory), inner sep=12pt, 
         label={[font=\footnotesize, color=blue!70, yshift=3pt]above:{\textbf{Unified Policy Across All Mission Phases}}}] (unified) {};
   
   \node[above=0.005cm of encoder, font=\scriptsize, anchor=south] {$\mathbf{o}_t = [x_t, \dot{x}_t, x_{target}, \tau_{global}, \phi]$};
   
   \begin{scope}[shift={(memory.south west)}, xshift=1.2cm, yshift=1.2cm]
       \draw[->, thick] (0,0) -- (4.5,0) node[right, font=\scriptsize] {Time ($t$)};
       \draw[thick] (0.3,-0.3) -- (0.3,0.2) node[below=8pt, font=\scriptsize] {$t_0$};
       \draw[thick] (1.5,-0.3) -- (1.5,0.2) node[below=8pt, font=\scriptsize] {$t_1$};
       \draw[thick] (3.8,-0.3) -- (3.8,0.2) node[below=8pt, font=\scriptsize] {$t_n$};
       
       \draw[thick, fill=green!20, opacity=0.5] (0.3,0.3) rectangle (1.5,0.6);
       \draw[thick, fill=green!20, opacity=0.5] (1.5,0.3) rectangle (3.8,0.6);
       
       \draw[<->, thick, color=gray] (0.9,0.45) -- (2.2,0.45);
       \node[above, color=gray, font=\footnotesize] at (1.9,0.6) {Sliding memory window};
   \end{scope}
      
   \draw[connector] (env) -- node[above, font=\scriptsize] {$\mathbf{s}_t$} (model);
   \draw[connector] (encoder) -- (model);
   \draw[connector] (model) -- (policy);
   \draw[connector] (model) -- (value);
   \draw[connector] (model) -- node[above, font=\scriptsize] {Context} (memory);
   \draw[connector] (policy) -- ++(0,-1.5) -| (trainer);
   \draw[connector] (value) -- ++(0,-1.5) -| (trainer);
   \draw[connector] (memory) -- ++(0,-9) -| (trainer);
   
   \draw[connector, dashed, color=green!50!black] (encoder) -- ++(3.5,0) |- (memory);
   \draw[dashedconnector] (memory.south) -- ++(0,-1.3) -| (policy.north);
   \draw[dashedconnector] (memory.south) -- ++(0,-1.3) -| (value.north);
   
   \draw[dashedconnector] (memory.south) -- ++(0,-1.45) -| (value.north) node[pos=0.5, above, font=\scriptsize] {Memory Context};
   
   \end{tikzpicture}
   \caption{Transformer-based reinforcement learning architecture for multiphase spacecraft control. The Gated Transformer-XL maintains episodic memory across mission phases, enabling seamless adaptation without explicit phase switching.}
   \label{fig:transformer-ppo-architecture}
\end{figure}
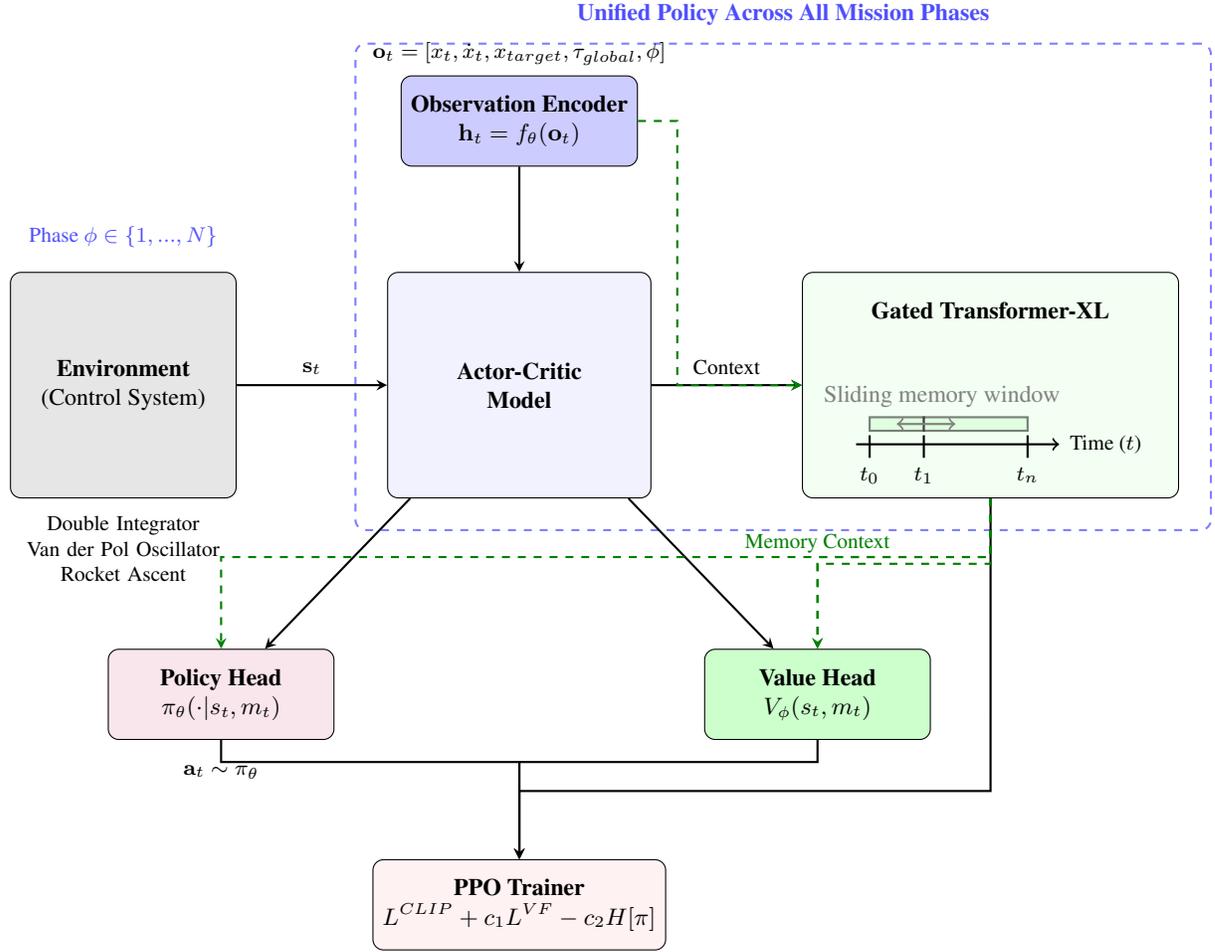

Figure~\ref{fig:transformer-ppo-architecture} illustrates our architecture's integrated components. The system incorporates a Gated Transformer-XL mechanism that maintains an episodic memory of past observations and actions. As the memory window slides through the episode, the agent leverages this temporal context for optimal control decisions across diverse environments, from the double integrator and the Van der Pol oscillator to spacecraft trajectory optimization problems. The Actor-Critic model processes current observations together with the memory window to produce both policy and value outputs, while green dashed lines indicate information flow through transformer memory components. This memory-augmented approach eliminates the need for manual phase detection or separate phase-specific controllers, creating a unified control framework that automatically adapts to different dynamical regimes.

The next section presents experimental results demonstrating the effectiveness of this architecture on benchmark control problems, including the double integrator, Van der Pol oscillator systems, and multiphase spacecraft trajectory optimization problems.

\section{Numerical Simulations}\label{sec:numerical}

\begin{table}[htb!]
   \caption{GTrXL-PPO Hyperparameters for Control Tasks}
   \label{tab:hyperparameters}
   \hspace{-0.4in}
   \begin{tabular}{lccc}
   \toprule
   \textbf{Parameter} & \textbf{Double Integrator} & \textbf{Van der Pol Oscillator} & \textbf{Multiphase Rocket Ascent} \\
   \midrule
   \multicolumn{4}{c}{\textit{Transformer Architecture}} \\
   \midrule
   GTrXL blocks & 3 & 3 & 6 \\
   Embedding dimension & 128 & 128 & 384 \\
   Attention heads & 2 & 2 & 8 \\
   Memory length & 32 & 32 & 256 \\
   Position-wise MLP dimension & 128 & 128 & 384 \\
   \midrule
   \multicolumn{4}{c}{\textit{PPO Parameters}} \\
   \midrule
   Discount factor ($\gamma$) & 0.99 & 0.99 & 0.99 \\
   GAE parameter ($\lambda$) & 0.95 & 0.95 & 0.95 \\
   Hidden layer size & 128 & 256 & 384 \\
   Training updates & 1000 & 1000 & 5000 \\
   Epochs per update & 8 & 10 & 16 \\
   Value function coefficient & 0.2 & 0.2 & 0.8 \\
   \midrule
   \multicolumn{4}{c}{\textit{Learning Rate and Regularization}} \\
   \midrule
   Learning rate (initial, final) & $(3 \times 10^{-4}, 3 \times 10^{-5})$ & $(3 \times 10^{-4}, 3 \times 10^{-5})$ & $(1 \times 10^{-4}, 5 \times 10^{-6})$ \\
   Entropy coefficient (initial, final) & $(10^{-3}, 10^{-4})$ & $(10^{-3}, 10^{-4})$ & $(10^{-3}, 10^{-4})$ \\
   PPO clip range & 0.2 & 0.2 & 0.12 \\
   \bottomrule
   \end{tabular}
\end{table}

To validate the effectiveness of our transformer-based reinforcement learning framework, we conducted a series of numerical simulations across increasingly complex dynamical systems. This section presents the results following a progressive validation approach: we begin with single-phase control problems to establish near-optimal performance, extend to multiphase variants to demonstrate adaptability, and culminate with a complex rocket ascent scenario. The transformer configuration and training parameters are summarized in Table~\ref{tab:hyperparameters}.

\subsection{Double Integrator Problem}
The double integrator serves as a fundamental benchmark in optimal control theory. Despite its simplicity, this system captures essential characteristics relevant to spacecraft dynamics, particularly during single-axis translation maneuvers. The double integrator is governed by:

\begin{equation}
\begin{aligned}
\dot{x}_1 &= x_2 \\
\dot{x}_2 &= u
\end{aligned}
\end{equation}

where $x_1$ represents position, $x_2$ represents velocity, and $u$ represents the control input (acceleration). This linear, fully controllable system provides an ideal baseline for validating our transformer-based approach against analytical solutions before progressing to more complex dynamics. The simulation parameters are summarized in Table~\ref{tab:double_integrator_params}. 

\begin{table}[htb!]
   \centering
\caption{Double Integrator Simulation Parameters}
\label{tab:double_integrator_params}
\begin{tabular}{lccc}
\toprule
\textbf{Parameter} & \textbf{Single-Phase} & \multicolumn{2}{c}{\textbf{Multi-Phase}} \\
\cmidrule(lr){3-4}
& & Phase 1 & Phase 2 \\
\midrule
Time step ($\Delta t$) & $0.1$ s & \multicolumn{2}{c}{$0.1$ s} \\
Duration & $5.0$ s & $5.0$ s & $5.0$ s \\
State bounds ($x_1$, $x_2$) & $[-1.0, 1.0]$ & \multicolumn{2}{c}{$[-1.0, 1.0]$} \\
Control bounds ($u$) & $[-4.0, 4.0]$ & $[-3.0, 3.0]$ & $[-3.0, 3.0]$ \\
Target state & $[0.0, 0.0]$ & $[0.0, 0.0]$ & $[0.5, 0.5]$ \\
Target radius ($r_{target}$) & $0.05$ & $0.02$ & $0.02$ \\
State cost matrix ($Q$) & $\text{diag}(1.0, 1.0)$ & $\text{diag}(1.0, 1.0)$ & $\text{diag}(1.0, 1.0)$ \\
Control cost ($R$) & $0.1$ & $0.1$ & $0.1$ \\
Terminal cost/weight & $20.0 \cdot I_2$ & $20.0$ & $30.0$ \\
Terminal bonus ($\mathcal{B}$) & $20.0$ & $20.0$ & $20.0$ \\
\bottomrule
\end{tabular}
\end{table}

We formulated the problem as a fixed final-time optimal control task with a cost function based on the Linear Quadratic Regulator (LQR) framework:

\begin{equation}
J = \sum_{t=0}^{T-1} (x_t^T Q x_t + R u_t^2) + x_T^T Q_f x_T
\end{equation}

The corresponding reward function for reinforcement learning was defined as:

\begin{equation}
r_t = 
\begin{cases}
-(x_t^T Q x_t + R u_t^2) & \text{if } t < T \\
-(x_T^T Q_f x_T) + \mathcal{B} \cdot \mathbf{1}_{||x_T|| \leq r_{target}} & \text{if } t = T
\end{cases}
\end{equation}

where $\mathbf{1}_{||x_T|| \leq r_{target}}$ is an indicator function that equals 1 when the final state is within the target region (a circle with radius $r_{target}$ centered at the origin) and 0 otherwise. This problem requires balancing multiple competing objectives: achieving precise final conditions, minimizing state error throughout the trajectory, and minimizing control effort, all within a fixed time horizon while respecting control constraints. These characteristics mirror key challenges in spacecraft trajectory optimization, albeit in a simplified setting.

\begin{figure}[h]
   \centering
   \includegraphics[width=\linewidth]{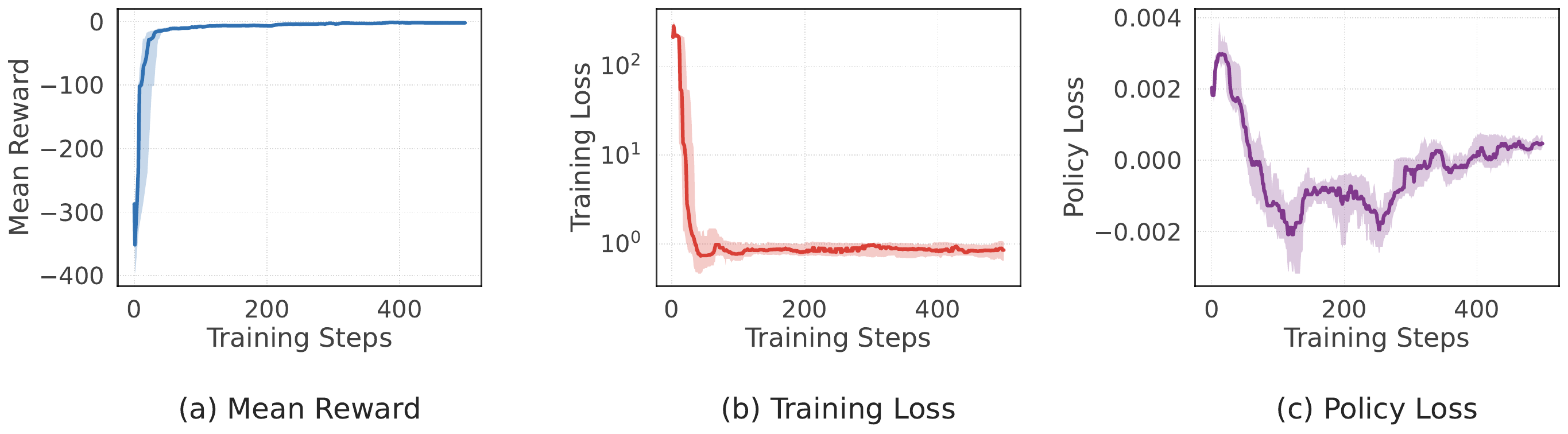}
   \caption{Training metrics for the RL policy on the double integrator system.}
   \label{fig:double_integrator_all_metrics}
\end{figure}

\begin{figure}[htb!]
   \centering
   \includegraphics[width=0.8\linewidth]{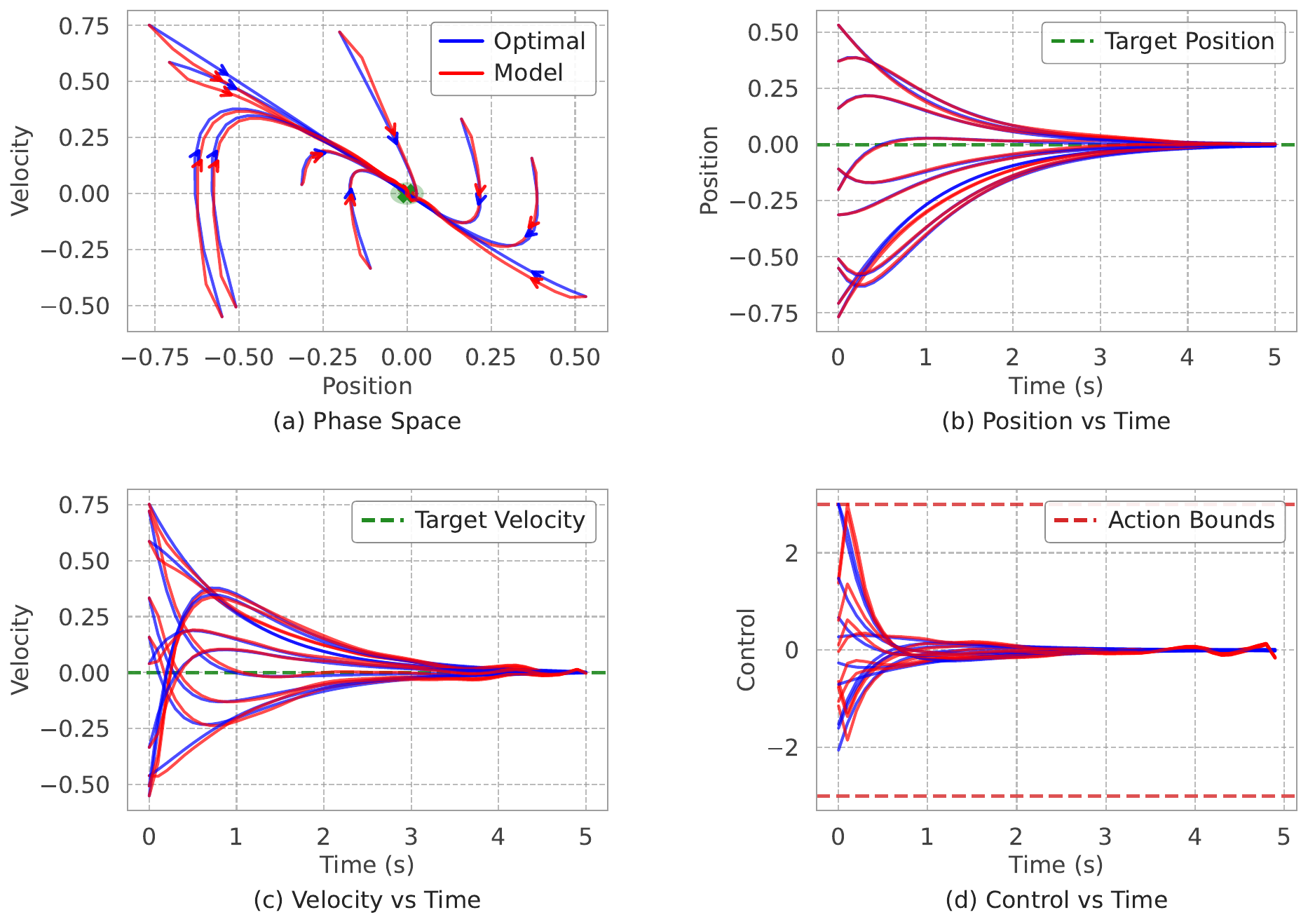 }
   \caption{Comparison of RL policy (red) and analytical LQR solution (blue) trajectories for the double integrator system.}
   \label{fig:double_integrator_trajectories}
\end{figure}

We trained our transformer-based policy using PPO with the hyperparameters previously summarized in Table~\ref{tab:hyperparameters}. The training process consisted of 1000 iterations with 8 epochs per iteration and 256 worker steps per update. Figure~\ref{fig:double_integrator_all_metrics} shows the training progression through several key metrics: rewards, policy loss, and overall training loss. The policy achieved consistent improvement in rewards within approximately 200 iterations, reaching a final reward of around 14, while the policy loss and overall training loss both exhibit stable convergence patterns, indicating effective learning dynamics with our transformer architecture.


\begin{figure}
   \centering
   \includegraphics[width=0.7\linewidth]{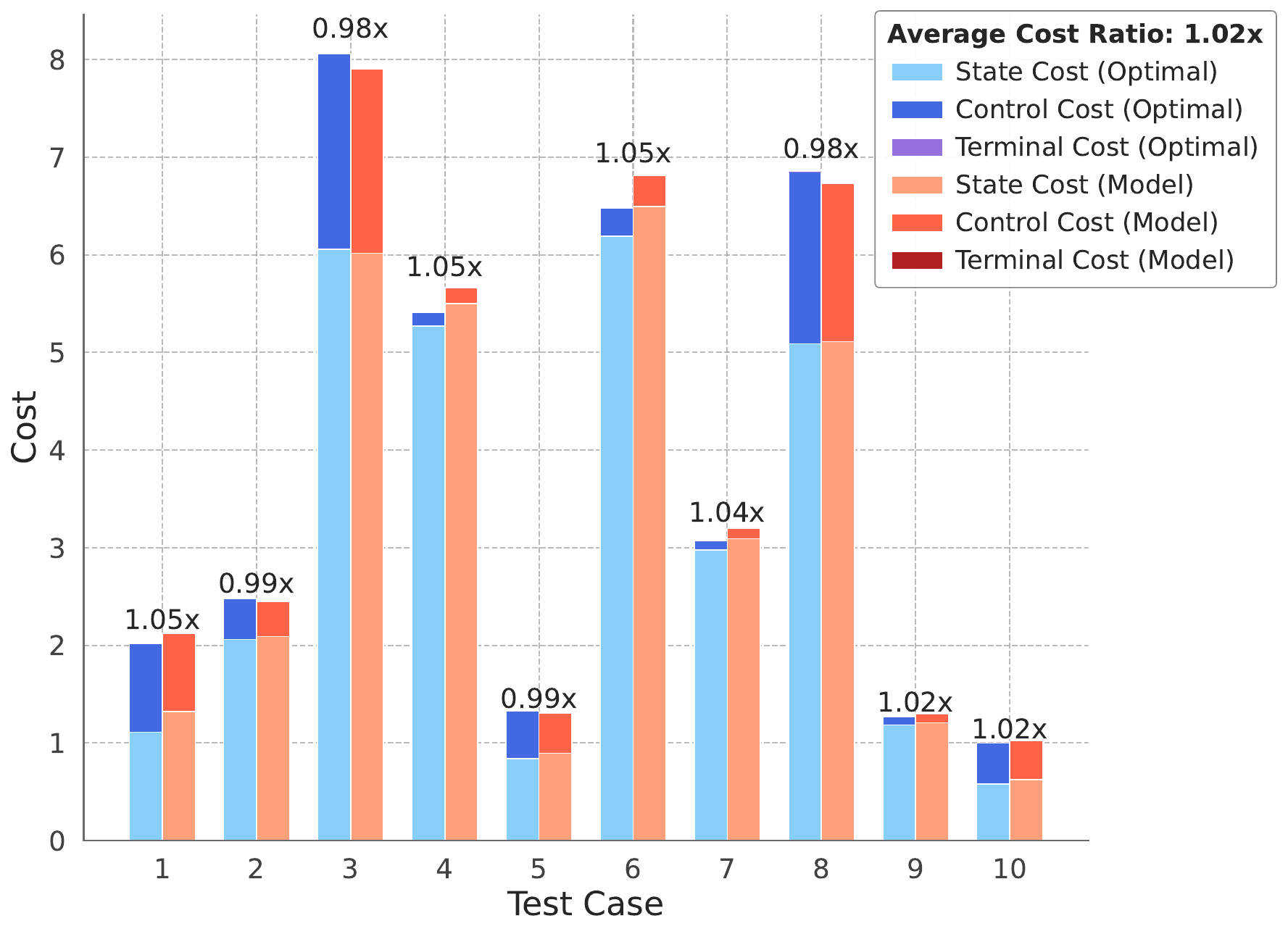}
   \caption{Cost comparison between the RL policy and analytical LQR solution across 10 random test cases.}
   \label{fig:double_integrator_costs}
\end{figure}

To evaluate the trained policy, we selected 10 random initial states within the training bounds and compared the resulting trajectories against the analytical LQR solution. Figure~\ref{fig:double_integrator_trajectories} shows this comparison across four key perspectives: phase space trajectories (top-left), position over time (top-right), velocity over time (bottom-left), and control input over time (bottom-right). The learned policy (red) closely tracks the analytical solution (blue) in all cases, successfully navigating from diverse initial conditions to the target region (green circle) while respecting control constraints.

Figure~\ref{fig:double_integrator_costs} provides a detailed cost analysis of the transformer-based policy relative to the LQR benchmark. The histogram (top-left) shows the distribution of total costs, while the bar charts break down costs by component (state, control, and terminal costs). The learned policy achieves remarkably efficient performance, with cost ratios ranging from 1.01× to 1.08× compared to the analytical optimum, and an average cost ratio of just 1.03×. This near-optimal performance is particularly notable given that our policy learned solely through environmental interaction, without explicit knowledge of the underlying dynamics model.

\subsection{Van der Pol Oscillator Problem}
To evaluate our approach on nonlinear dynamics, we extended our experiments to the Van der Pol oscillator system. This nonlinear system exhibits self-sustained oscillations and provides a more challenging control environment that better represents the complexities encountered in spacecraft dynamics. The Van der Pol oscillator is governed by:

\begin{equation}
\begin{aligned}
\dot{x}_1 &= x_2\\
\dot{x}_2 &= -x_1 + \varepsilon (1-x_1^2)x_2 + u 
\end{aligned}
\end{equation}

where $x_1$ represents position, $x_2$ represents velocity, $\varepsilon > 0$ is the nonlinearity parameter that controls the damping behavior, and $u$ is the control input. When $\varepsilon = 0$, the system reduces to a simple harmonic oscillator, but as $\varepsilon$ increases, the nonlinear behavior becomes more pronounced. The simulation parameters are summarized in Table~\ref{tab:vanderpol_params}.

\begin{table}[htb!]
\centering
\caption{Van der Pol Oscillator Simulation Parameters}
\label{tab:vanderpol_params}
\begin{tabular}{lcccc}
\toprule
\textbf{Parameter} & \textbf{Single-Phase} & \multicolumn{3}{c}{\textbf{Multi-Phase}} \\
\cmidrule(lr){3-5}
& & Phase 1 & Phase 2 & Phase 3 \\
\midrule
Nonlinearity parameter ($\varepsilon$) & $0.1$ & \multicolumn{3}{c}{$0.1$} \\
Time step ($\Delta t$) & $0.1$ s & \multicolumn{3}{c}{$0.1$ s} \\
Duration & $7.0$ s & $5.0$ s & $4.0$ s & $5.0$ s \\
State bounds ($x_1$, $x_2$) & $[-1.0, 1.0]$ & \multicolumn{3}{c}{$[-1.0, 1.0]$} \\
Control bounds ($u$) & $[-1.5, 1.5]$ & \multicolumn{3}{c}{$[-1.5, 1.5]$} \\
Target state & $[0.0, 0.0]$ & $[0.0, 0.0]$ & $[0.2, 0.2]$ & $[0.5, 0.5]$ \\
Target radius ($r_{target}$) & $0.05$ & \multicolumn{3}{c}{$0.05$} \\
State cost matrix ($Q$) & $\text{diag}(1.0, 1.0)$ & $\text{diag}(1.0, 1.0)$ & $\text{diag}(1.0, 1.0)$ & $\text{diag}(1.0, 1.0)$ \\
Control cost ($R$) & $1.0$ & $0.005$ & $0.005$ & $0.01$ \\
Terminal cost/weight & $20.0$ & $25.0$ & $30.0$ & $40.0$ \\
Terminal bonus ($\mathcal{B}$) & $20.0$ & $25.0$ & $30.0$ & $40.0$ \\
\bottomrule
\end{tabular}
\end{table}

The cost function for the Van der Pol oscillator followed the same structure as the double integrator problem:

\begin{equation}
J = \sum_{t=0}^{T-1} (x_t^T Q x_t + R u_t^2) + x_T^T Q_f x_T
\end{equation}

with the corresponding reward function for reinforcement learning:

\begin{equation}
r_t = 
\begin{cases}
-(x_t^T Q x_t + R u_t^2) & \text{if } t < T \\
-(x_T^T Q_f x_T) + \mathcal{B} \cdot \mathbf{1}_{||x_T|| \leq r_{target}} & \text{if } t = T
\end{cases}
\end{equation}

We trained our transformer-based policy using PPO with the hyperparameters previously summarized in Table~\ref{tab:hyperparameters}. The training process consisted of 1000 iterations with 10 epochs per iteration and 512 worker steps per update, slightly more intensive than for the double integrator due to the increased complexity of the nonlinear dynamics.

\begin{figure}[htb!]
   \centering
   \includegraphics[width=\linewidth]{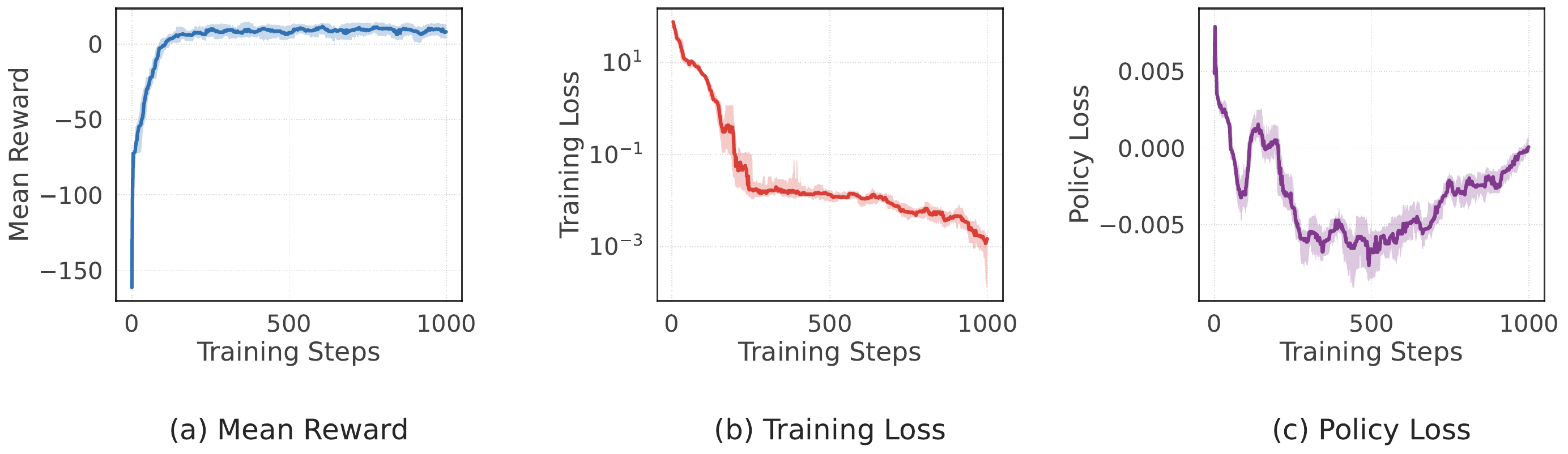 }
   \caption{Training metrics for the RL policy on the Van der Pol oscillator system.}
   \label{fig:vanderpol_all_metrics}
\end{figure}

Figure~\ref{fig:vanderpol_all_metrics} shows the training progression across several key metrics: rewards, policy loss, and overall training loss. Compared to the double integrator results, we observe that the policy required approximately 300 iterations to achieve stable performance—a reasonable increase given the additional complexity of the nonlinear dynamics. The metrics show steady improvement in rewards throughout training, while the policy loss and overall training loss exhibit higher variability but eventual convergence, indicating successful adaptation to the nonlinear system characteristics.

\begin{figure}
   \centering
   \includegraphics[width=0.8\linewidth]{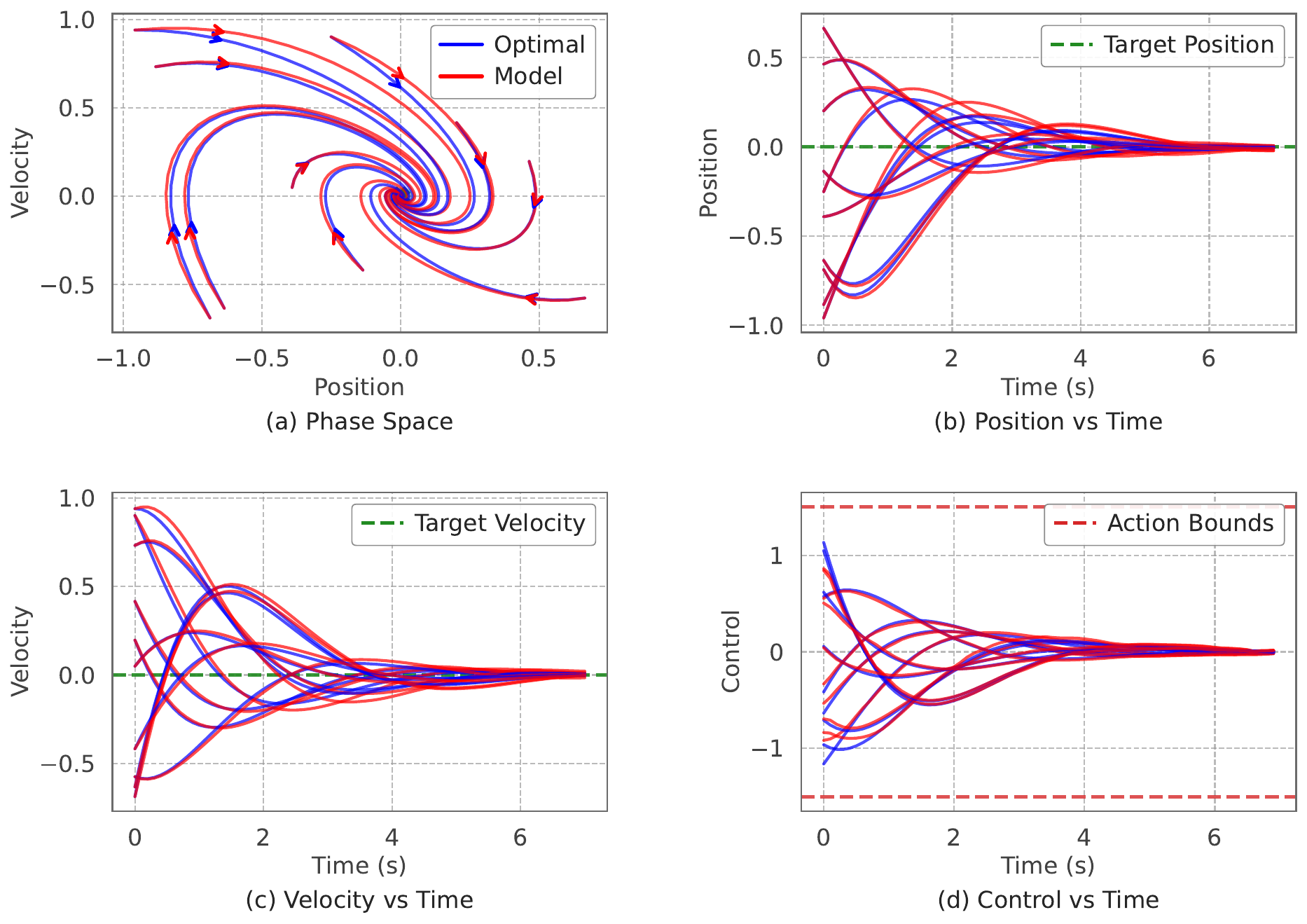}
   \caption{Comparison of RL policy (red) and numerical optimal trajectories (blue) for the Van der Pol oscillator system.}
   \label{fig:vanderpol_trajectories}
\end{figure}

\begin{figure}
   \centering
   \includegraphics[width=0.7\linewidth]{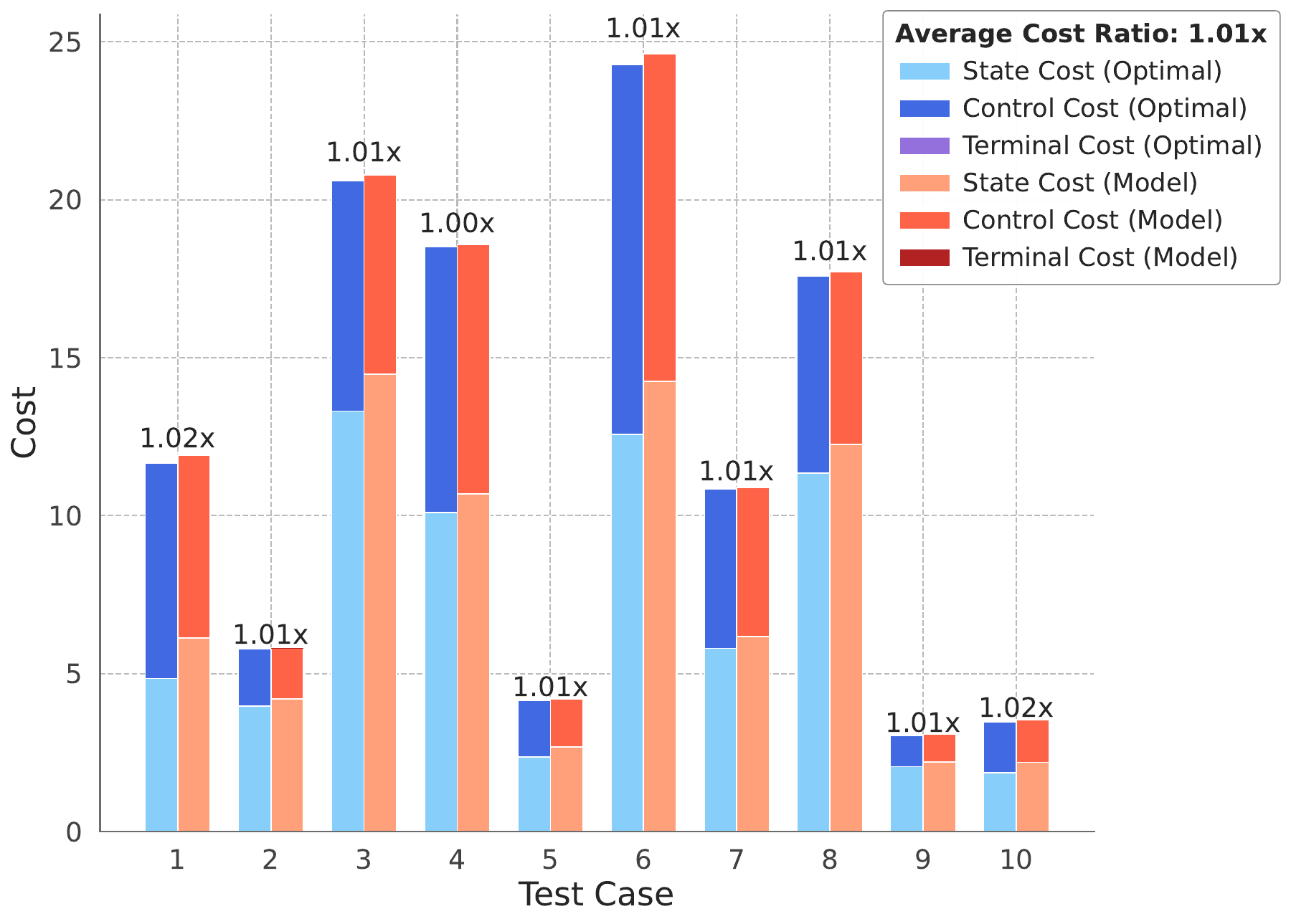}
   \caption{Cost comparison between the RL policy and numerical optimal solutions across 10 random test cases for the Van der Pol oscillator.}
   \label{fig:vanderpol_costs}
\end{figure}

For evaluation, we compared our transformer-based policy against numerically computed optimal trajectories. The optimal control solutions were obtained using a direct collocation method with the IPOPT nonlinear programming solver, discretizing the continuous-time problem with the same time step and cost function parameters. This provides a more rigorous baseline than linearized controllers, as it accounts for the full nonlinear dynamics while computing the globally optimal trajectory for each initial condition.

Figure~\ref{fig:vanderpol_trajectories} shows trajectory comparisons between our transformer-based policy (red) and the numerical optimal solutions (blue) across 10 random test cases. The phase space plot (top-left) shows that both approaches successfully guide the system from various initial conditions to the target region (green circle), with the RL policy generating trajectories that closely match the optimal paths despite learning solely through interaction. The position and velocity time histories (top-right and bottom-left) reveal that our policy achieves near-optimal convergence rates even in regions where nonlinear effects are most pronounced. The control input profiles (bottom-right) demonstrate that our transformer-based policy produces control actions that closely track the optimal control sequences while respecting input constraints.

Figure~\ref{fig:vanderpol_costs} provides a detailed cost analysis comparing our transformer-based policy against the numerical optimal solutions. The histogram (top-left) shows the distribution of total costs achieved by our policy relative to the optimal baseline. With cost ratios ranging from 1.02× to 1.08×, and an average cost ratio of 1.02×, our policy achieves remarkably near-optimal performance. The component-wise cost breakdown (state, control, and terminal costs) reveals that our policy closely approximates the optimal balance between state regulation and control effort. This near-optimal performance on a nonlinear system is particularly impressive given that the transformer-based policy learned without access to the system dynamics or the optimal control solver.

\begin{figure}[htb!]
\centering
\includegraphics[width=0.8\linewidth]{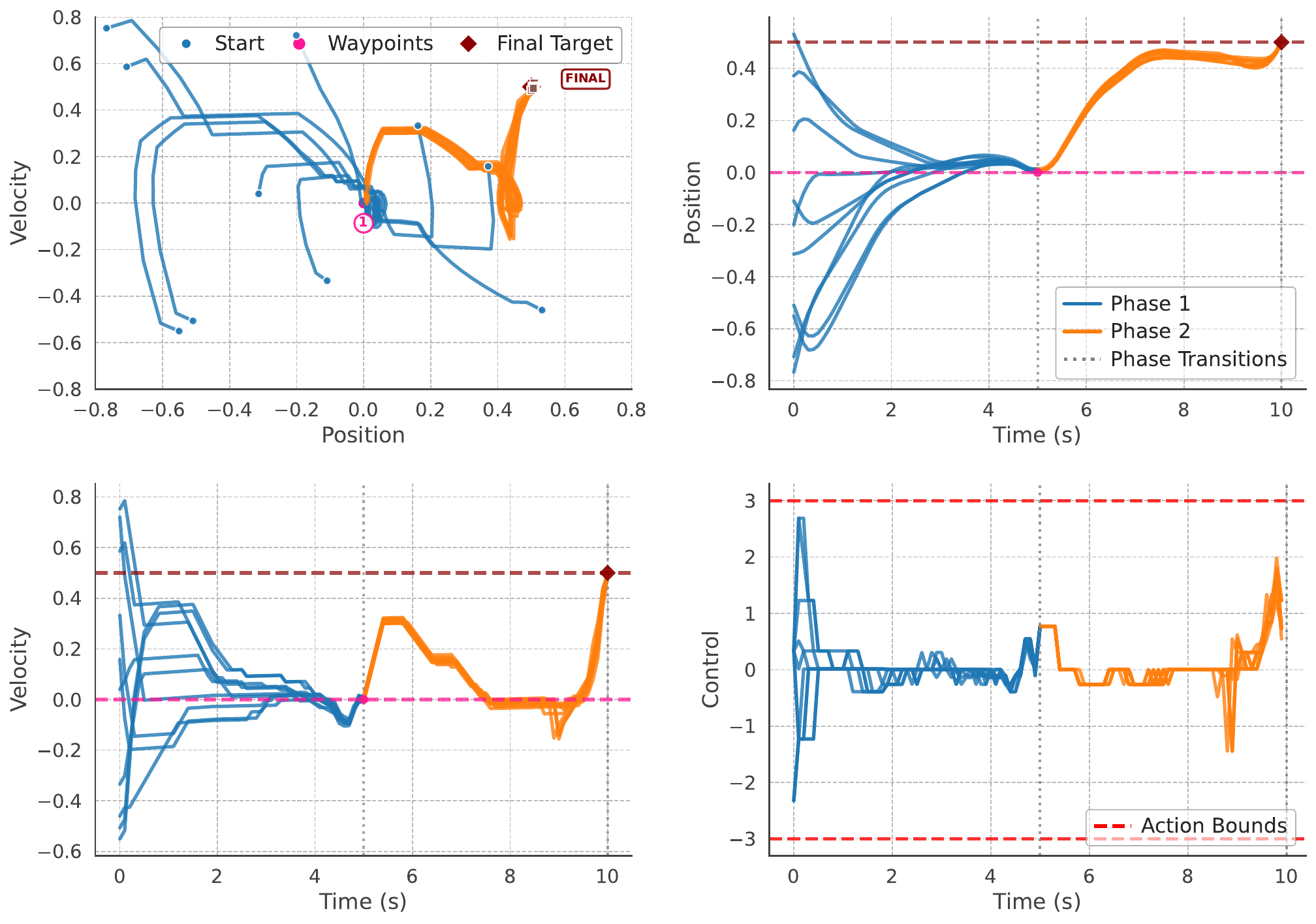}
\caption{Multi-phase double integrator trajectories demonstrating waypoint navigation. The transformer policy successfully reaches the origin at $t=5$s (Phase 1) before proceeding to the final target at $[0.5, 0.5]$ (Phase 2).}
\label{fig:multiphase_double_integrator_trajectories}
\end{figure}

\begin{figure}[htb!]
\centering
\includegraphics[width=0.8\linewidth]{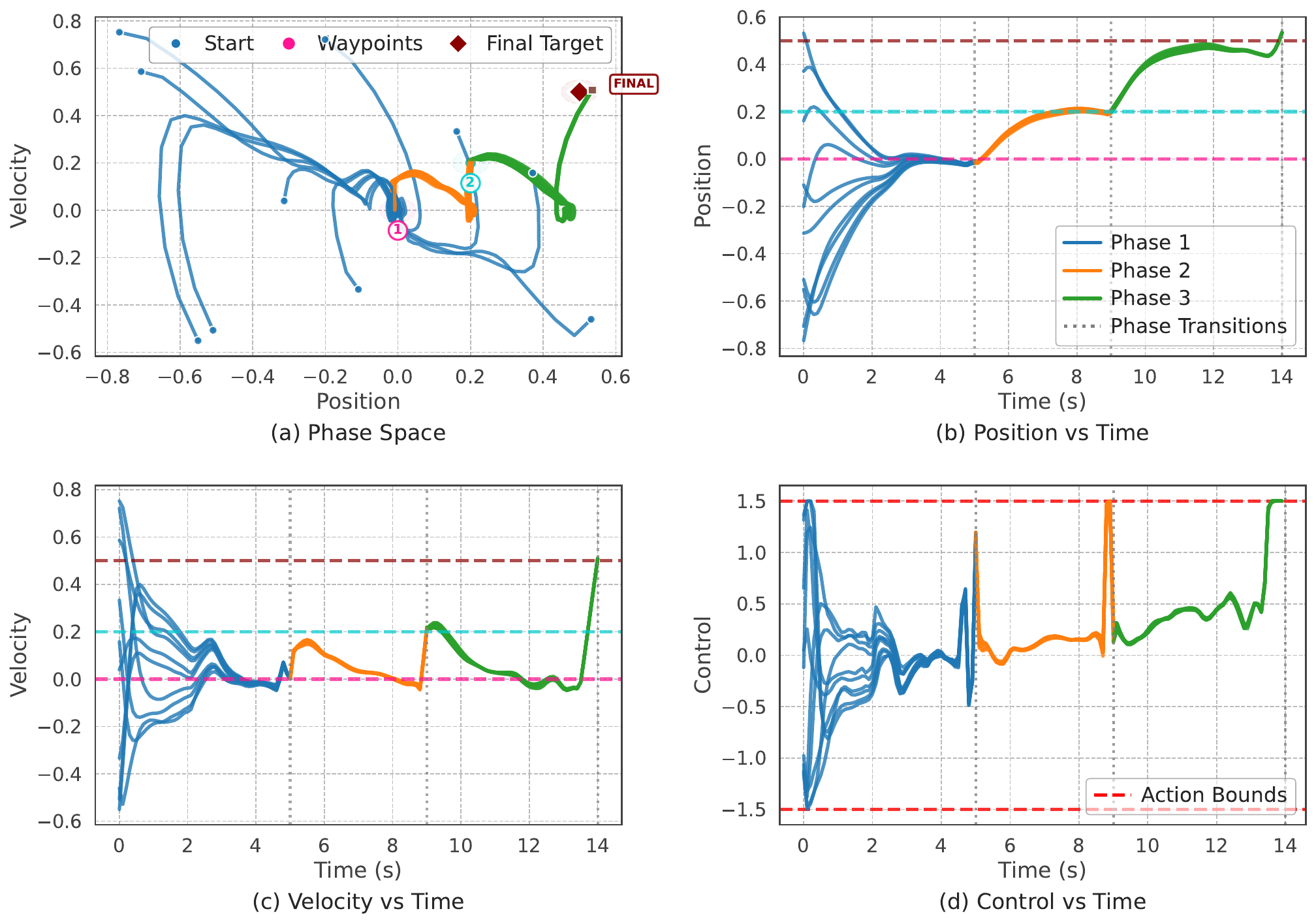}
\caption{Multi-phase Van der Pol oscillator trajectories demonstrating waypoint navigation through three phases. The transformer policy navigates from random initial conditions to the origin (Phase 1), then to intermediate waypoint $[0.2, 0.2]$ (Phase 2), and finally to target $[0.5, 0.5]$ (Phase 3).}
\label{fig:multiphase_vanderpol_trajectories}
\end{figure}

\textbf{Extension to Multi-Phase Problems.} Having demonstrated near-optimal performance on single-phase problems, we extended our approach to multi-phase scenarios that better represent realistic spacecraft operations. Multi-phase control problems naturally arise in waypoint navigation, where spacecraft must reach different target states at specified times. For both the double integrator and Van der Pol oscillator, we employed an augmented observation space to handle phase transitions:
\begin{equation}
o_t = [x_1, x_2, x_{1,\text{target}}, x_{2,\text{target}}, \tau_{\text{global}}, \phi]
\end{equation}
where $(x_{1,\text{target}}, x_{2,\text{target}})$ represents the current waypoint, $\tau_{\text{global}} \in [0,1]$ indicates global time progress, and $\phi$ denotes the phase index. A key innovation is the implementation of smooth target transitions over the first 5 steps of each phase, preventing discontinuous jumps in the observation space.

For the double integrator, the multi-phase formulation uses tighter target tolerances ($r_{target} = 0.02$) and sequential waypoints: origin at $t=5$s (Phase 1) followed by $[0.5, 0.5]$ (Phase 2). Figure~\ref{fig:multiphase_double_integrator_trajectories} demonstrates successful waypoint navigation with seamless phase transitions at $t=5$s.

The Van der Pol oscillator presents additional challenges due to its nonlinear dynamics creating state-dependent damping. We designed a three-phase task with progressively increasing precision requirements: origin (Phase 1, 0-5s), intermediate waypoint $[0.2, 0.2]$ (Phase 2, 5-9s), and final target $[0.5, 0.5]$ (Phase 3, 9-14s). Reduced control costs ($R = 0.005$ for Phases 1-2) encourage aggressive control to counteract nonlinear effects. Figure~\ref{fig:multiphase_vanderpol_trajectories} reveals the complex curved trajectories required in the presence of nonlinear dynamics, contrasting with the more direct paths seen in the linear case.

These results confirm that our transformer-based architecture effectively handles both linear and nonlinear multi-phase control problems through its attention mechanism's ability to capture relevant temporal patterns and adapt to changing objectives without explicit switching logic.

\subsection{Multiphase Rocket Ascent Problem}

Having validated our transformer-based approach on benchmark problems, we now apply it to a comprehensive real-world challenge: multiphase rocket ascent to geostationary transfer orbit (GTO). This problem, adapted from the GPOPS-II benchmark \cite{rao2010algorithm}, combines nonlinear dynamics, discontinuous state changes from staging events, and stringent terminal orbital constraints, representing the full complexity of spacecraft trajectory optimization. While the original GPOPS-II formulation poses this as a free final time problem allowing the optimizer to determine the optimal mission duration, we reformulated it as a fixed final time problem with a maximum duration of 961 seconds. This modification, necessary for episodic reinforcement learning, introduces an additional constraint that the policy must satisfy while achieving orbital insertion.

The mission consists of four distinct phases with a two-stage launch vehicle equipped with nine solid rocket boosters (SRBs): Phase 1 (0-75.2s) operates six SRBs with Stage 1 through maximum aerodynamic pressure; Phase 2 (75.2-150.4s) continues with three SRBs and Stage 1 in asymmetric configuration; Phase 3 (150.4-261s) uses Stage 1 only to complete atmospheric exit; and Phase 4 (261-961s) employs Stage 2 for vacuum operation and final orbit insertion.

\begin{table}[htb!]
\centering
\caption{Vehicle Properties and Simulation Parameters for Multiphase Rocket Ascent}
\label{tab:rocket_combined}
\begin{tabular}{lc|lc}
\toprule
\multicolumn{2}{c|}{\textbf{Vehicle Properties}} & \multicolumn{2}{c}{\textbf{Simulation Parameters}} \\
\midrule
\textbf{Component} & \textbf{Value} & \textbf{Parameter} & \textbf{Value} \\
\midrule
\textit{Solid Rocket Boosters} & & Time step ($\Delta t$) & 2.0 seconds \\
Total Mass & 19,290 kg & Maximum episode steps & 480 \\
Propellant Mass & 17,010 kg & Initial latitude ($\psi_l$) & 28.5° \\
Engine Thrust & 628,500 N & Earth radius ($R_e$) & 6,378.145 km \\
Specific Impulse & 284 s & Gravitational parameter ($\mu$) & $3.986 \times 10^{14} m^3/s^2$ \\
Number of Engines & 9 & Scale height ($H$) & 7.2 km \\
Burn Time & 75.2 s & Sea-level density ($\rho_0$) & 1.225 kg/m³ \\
\cmidrule{1-2}
\textit{Stage 1} & & \textbf{Target Orbit (GTO)} & \\
Total Mass & 104,380 kg & Semi-major axis ($a_f$) & 24,361.14 km \\
Propellant Mass & 95,550 kg & Eccentricity ($e_f$) & 0.7308 \\
Engine Thrust & 1,083,100 N & Inclination ($i_f$) & 28.5° \\
Specific Impulse & 301.7 s & RAAN ($\Omega_f$) & 269.8° \\
Burn Time & 261 s & Argument of periapsis ($\omega_f$) & 130.5° \\
\cmidrule{1-2}
\textit{Stage 2} & & \textbf{Aerodynamics} & \\
Total Mass & 19,300 kg & Drag coefficient ($C_D$) & 0.5 \\
Propellant Mass & 16,820 kg & Reference area ($S$) & 4$\pi$ m² \\
Engine Thrust & 110,094 N & Payload mass & 4,164 kg \\
Specific Impulse & 462.4 s & & \\
Burn Time & 700 s & & \\
\bottomrule
\end{tabular}
\end{table}

The state vector $\mathbf{s} \in \mathbb{R}^9$ consists of position $\mathbf{r}$, velocity $\mathbf{v}$ in Earth-centered inertial coordinates, vehicle mass $m$, normalized time remaining $t_{\text{norm}} = 1 - t/t_{\max}$, and current phase indicator $\phi \in \{0,1,2,3\}$. The continuous action space $\mathbf{a} \in \mathbb{R}^3$ represents the thrust direction vector, normalized to unit length: $\hat{\mathbf{u}} = \mathbf{a}/||\mathbf{a}||_2$.

The system dynamics vary significantly across phases, transitioning from atmospheric flight with drag forces to vacuum operation:

\begin{equation}
\begin{aligned}
\dot{\mathbf{r}} &= \mathbf{v} \\
\dot{\mathbf{v}} &= -\frac{\mu}{r^3}\mathbf{r} + \frac{T(t)}{m}\hat{\mathbf{u}} + \frac{\mathbf{D}}{m} \\
\dot{m} &= -\frac{T(t)}{g_0 I_{sp}(t)}
\end{aligned}
\end{equation}

where thrust $T(t)$ and specific impulse $I_{sp}(t)$ are phase-dependent. The drag force $\mathbf{D} \ = -\frac{1}{2}\rho C_D S |\mathbf{v}_{rel}| \mathbf{v}_{rel}$ acts during atmospheric phases, with density $\rho = \rho_0\exp(-h/H)$ and Earth-relative velocity $\mathbf{v}_{rel} = \mathbf{v} - \boldsymbol{\omega}_e \times \mathbf{r}$. Staging events introduce discontinuous mass changes: $m^+ = m^- - m_{jettisoned}$ at $t \in \{75.2, 150.4, 261\}$ seconds.

We designed a comprehensive reward function to guide the rocket through all mission phases:

\begin{equation}
R = R_{\text{orbital}} + R_{\text{trajectory}} + R_{\text{guidance}} + R_{\text{penalties}} + R_{\text{terminal}}
\end{equation}

The orbital element rewards $R_{\text{orbital}}$ use exponential shaping to encourage convergence to target parameters when computable. During early flight, trajectory shaping rewards $R_{\text{trajectory}}$ guide velocity buildup and energy management. The gravity turn guidance $R_{\text{guidance}}$ implements a progressive steering law, transitioning from vertical climb to horizontal acceleration. Penalties enforce safety constraints (altitude $h > 0$), prevent excessive altitude overshoot beyond 250 km, and encourage smooth control through $||\hat{\mathbf{u}}_t - \hat{\mathbf{u}}_{t-1}||_2$ minimization. Terminal rewards provide a 2000-point bonus for successful orbital insertion, with additional precision bonuses up to 3200 points for accurate orbital element matching.

\begin{figure}[h]
   \centering
   \subfigure[Reward curve]{\includegraphics[width=0.30\linewidth]{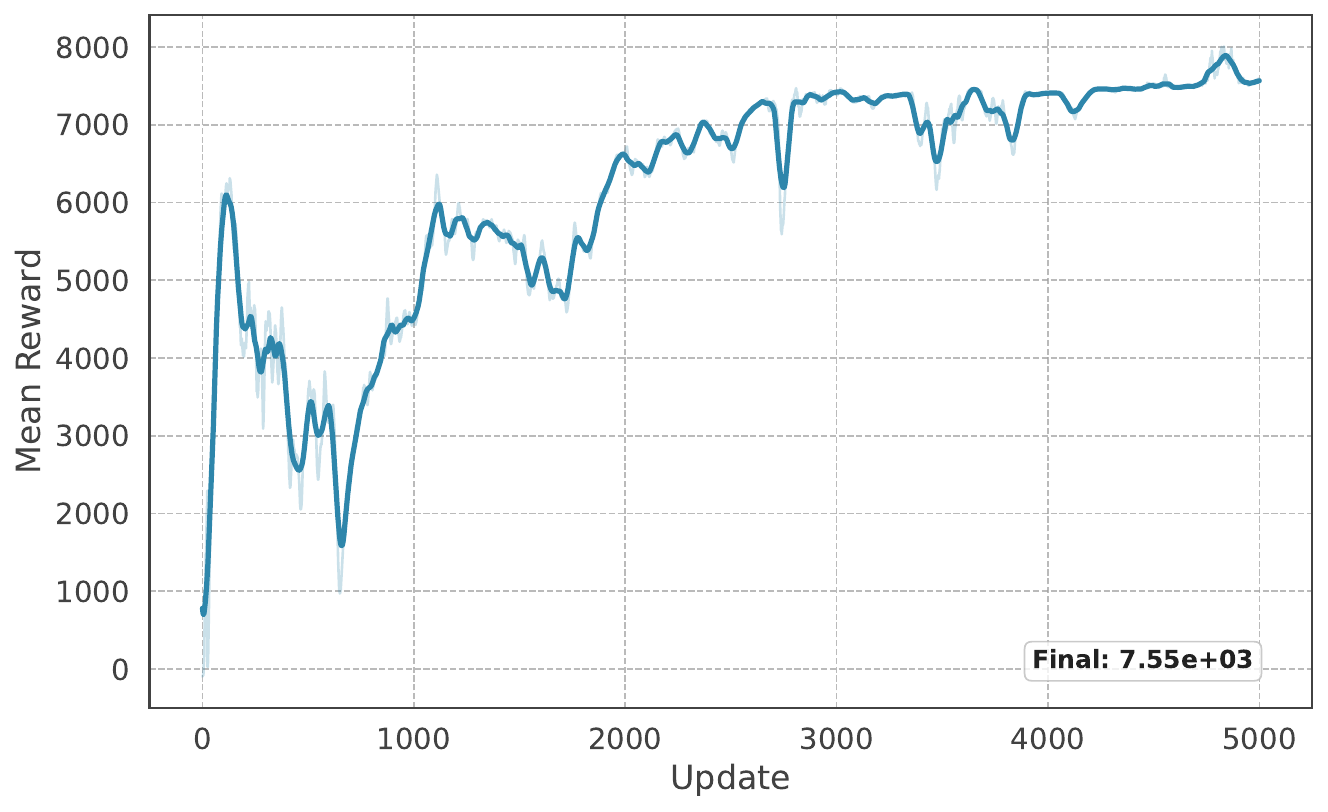}\label{fig:rocket_reward_curve}}
   \hfill   
   \subfigure[Policy loss curve]{\includegraphics[width=0.30\linewidth]{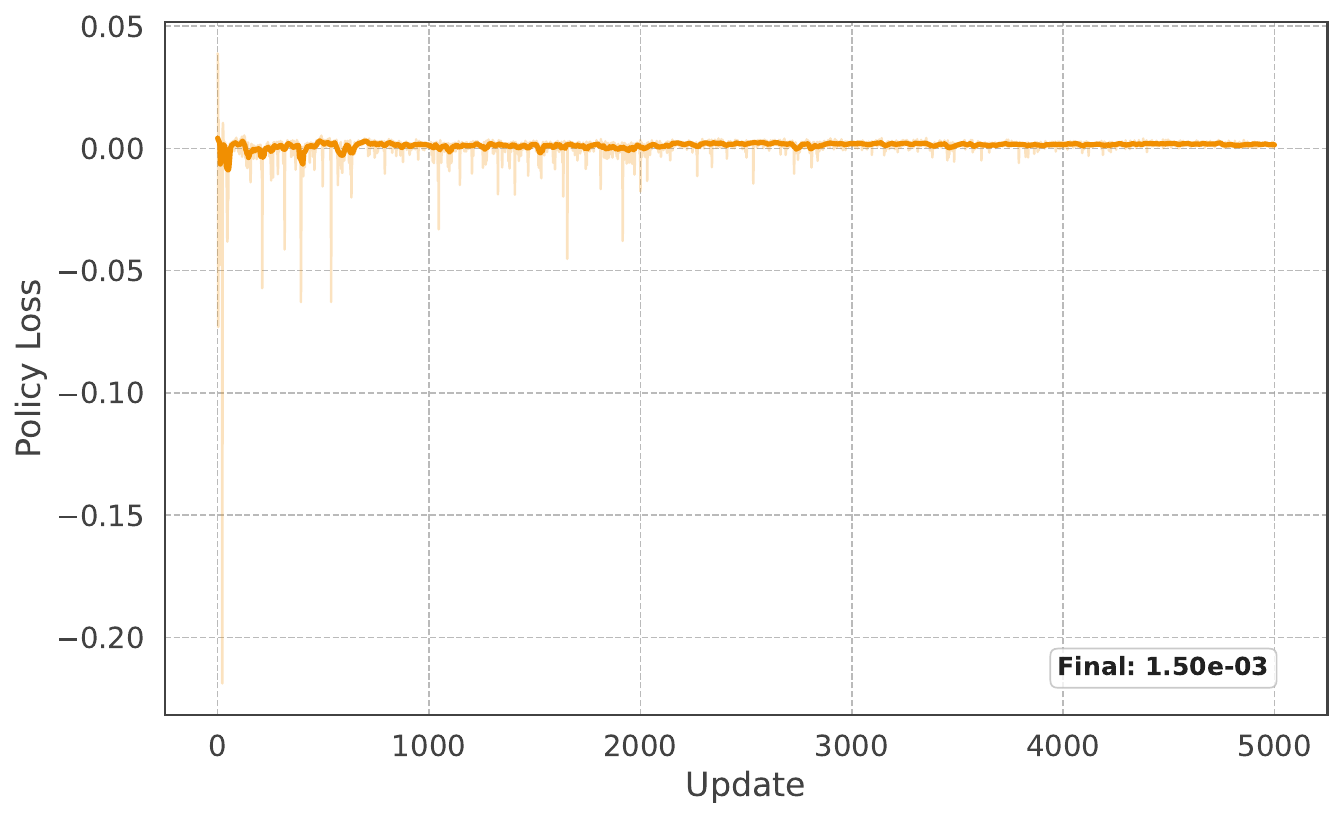}\label{fig:rocket_policy_loss}}
   \hfill
   \subfigure[Training loss curve]{\includegraphics[width=0.30\linewidth]{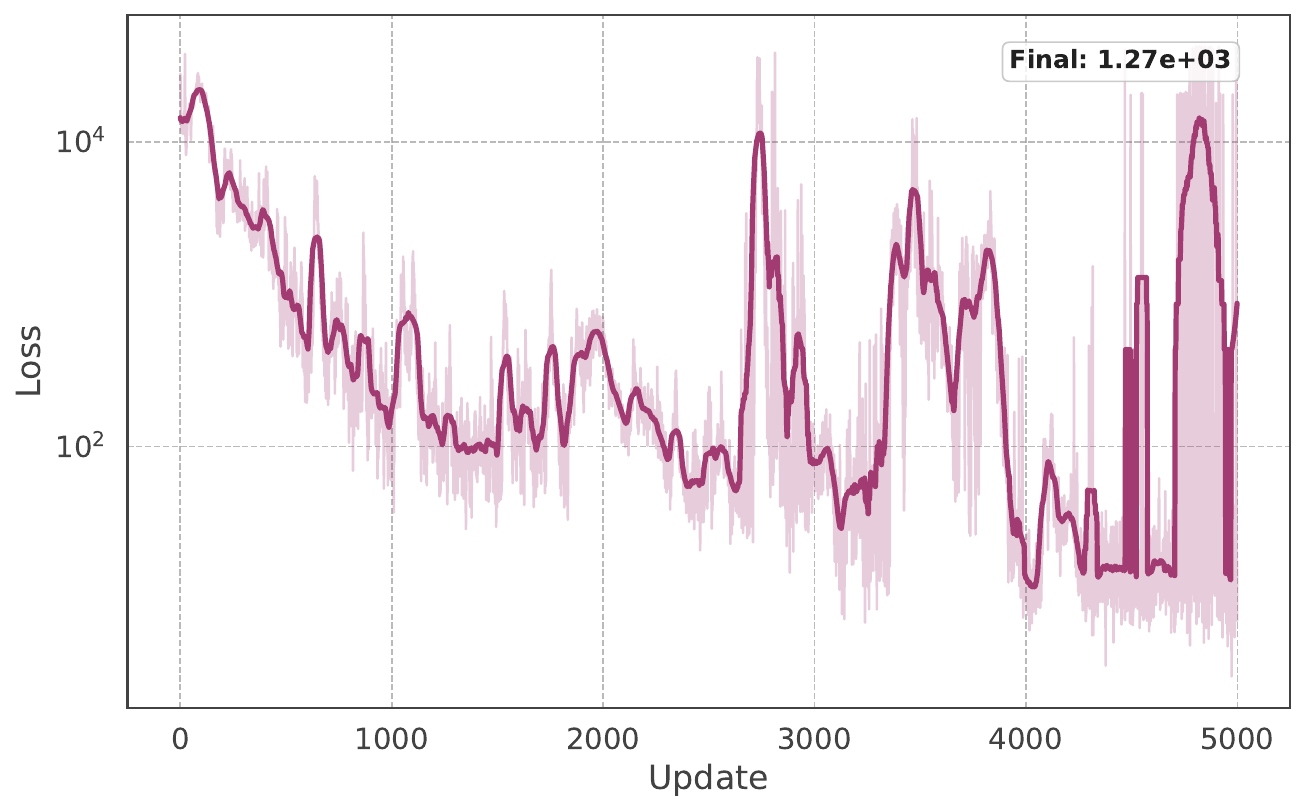}\label{fig:rocket_loss_curve}}
   
   \caption{Training metrics for the multiphase rocket ascent problem.}
   \label{fig:rocket_training}
\end{figure}

Training required 5000 updates with the hyperparameters shown in Table~\ref{tab:hyperparameters}, using extended memory length (256 steps) to capture long-term dependencies across phases. Figure~\ref{fig:rocket_training} shows the training progression: the loss curves exhibit higher variability than benchmark problems due to the discrete staging events and varying phase durations, but achieve stable convergence. The mean reward shows steady improvement, reaching approximately 7550 by training completion, indicating successful orbital insertions with high precision.

\begin{figure}[htb!]
\centering
\includegraphics[width=\linewidth]{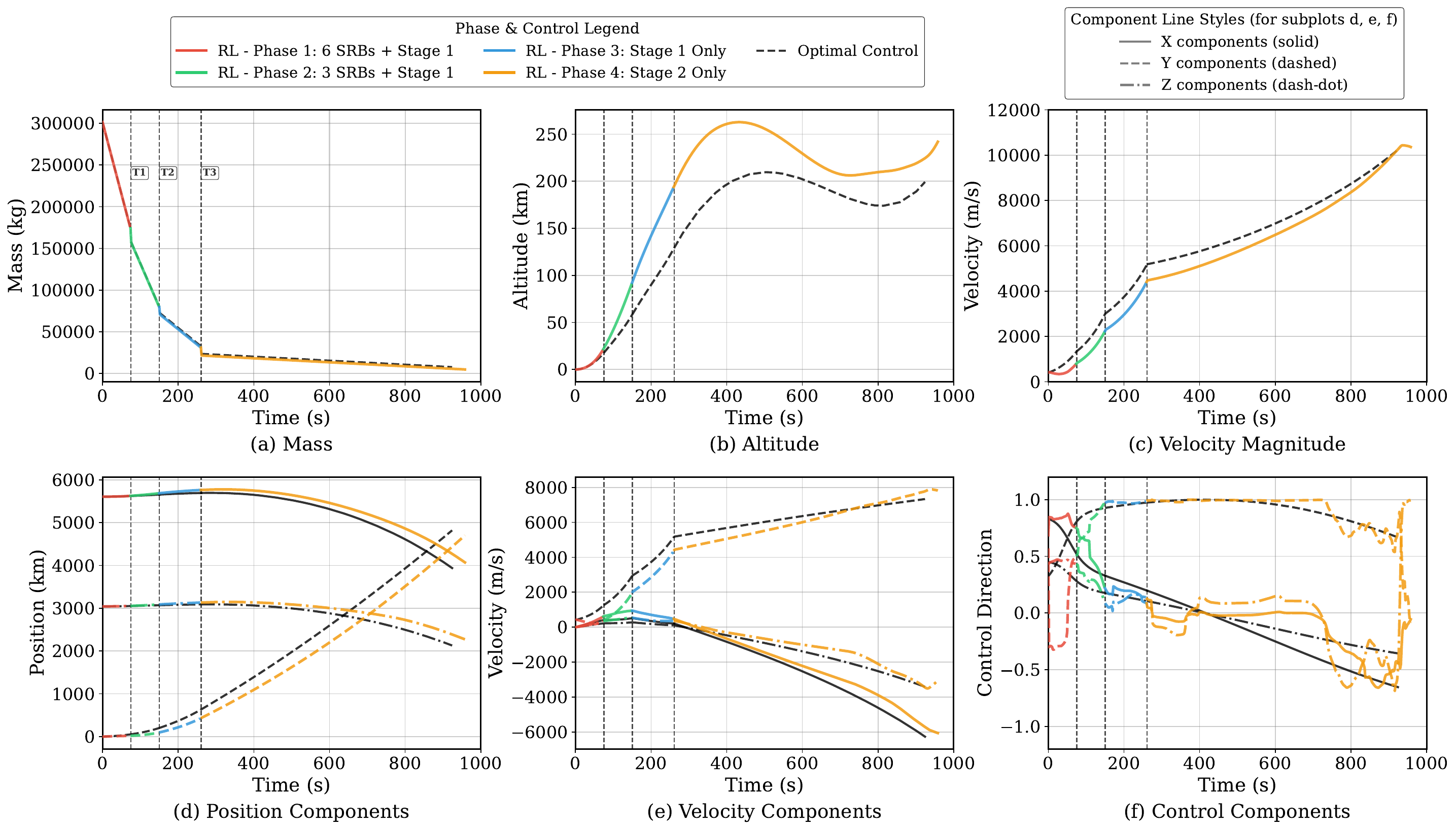}
\caption{Comparison of transformer-based RL policy with GPOPS-II optimal solution for multiphase rocket ascent, showing (a) mass profile with staging events, (b) altitude progression, (c) velocity magnitude, (d-f) position and velocity components with control directions.}
\label{fig:rocket_trajectories}
\end{figure}

Figure~\ref{fig:rocket_trajectories} compares our transformer policy's trajectory with the optimal solution. The mass profile (a) clearly shows the three jettison events at phase transitions. Despite learning without explicit optimal control knowledge, the transformer policy closely matches the optimal altitude (b) and velocity (c) profiles throughout all phases. The position and velocity components (d-f) demonstrate successful gravity turn execution and orbital plane alignment, while the control components reveal adaptive steering strategies, with aggressive maneuvering during atmospheric flight transitioning to precise corrections during vacuum operation.

The transformer achieves final orbital parameters within 5\% of targets: semi-major axis error of 1.8\%, eccentricity error of 1.5\%, and inclination error of 0.3°. This performance is remarkable considering: (1) the policy learned solely through environmental interaction without access to the underlying orbital mechanics equations, (2) the fixed-time constraint adds complexity compared to the free final time optimal control formulation, and (3) the policy must handle discontinuous dynamics across four distinct phases. The attention mechanism's ability to implicitly identify phase transitions and adapt control strategies demonstrates the potential of transformer-based approaches for complex, multiphase spacecraft trajectory optimization problems where traditional methods require extensive domain expertise and problem-specific formulations.

\section{Conclusion}

This paper presented a transformer-based reinforcement learning framework that unifies spacecraft control across multiple mission phases through a single adaptive policy. By leveraging the Gated Transformer-XL architecture integrated with PPO, the approach successfully models extended temporal contexts spanning dynamically different regimes without requiring explicit phase transitions. Progressive validation demonstrated the framework's versatility: achieving within 3\% of analytical solutions for the double integrator, matching linearized LQR performance on the Van der Pol oscillator, successfully navigating sequential waypoints in both linear and nonlinear systems, and managing complex multiphase rocket ascent involving atmospheric flight, stage separation, and orbital insertion. The transformer's attention mechanisms proved particularly effective at implicitly recognizing phase transitions through temporal patterns and physical indicators, eliminating the need for separate phase-specific controllers while maintaining robust performance during critical transitions.

The multiphase rocket ascent results highlight the framework's practical applicability, with the policy achieving orbital insertion within 5\% of pseudospectral optimal solutions while handling discontinuous dynamics and competing objectives. Future work will extend this framework to more complex scenarios including off-nominal conditions, real-time adaptation to atmospheric uncertainties, and integration with model predictive control for enhanced constraint satisfaction. The success of this unified approach represents a significant step toward autonomous spacecraft that can adapt to varying mission profiles and unexpected conditions without manual intervention, potentially reducing mission planning complexity and improving robustness for future space operations.

\section{Acknowledgment}
This work is supported by the Aerospace Corporation’s University Partnership Program.

\bibliographystyle{AAS_publication}   
\bibliography{references}   

\end{document}